\documentclass[11pt,letterpaper]{article}
\usepackage{emnlp2017}
\usepackage{times}
\usepackage{latexsym}
\usepackage{enumitem}

\usepackage{mathtools}
\usepackage{amsmath}
\usepackage{MnSymbol}
\usepackage{amsfonts}
\usepackage{xcolor}
\usepackage{colortbl}
\usepackage{graphicx}
\usepackage{subcaption}
\usepackage[font=small]{caption}
\usepackage{multirow}
\usepackage{microtype}
\usepackage{multicol}
\usepackage{tabularx} 
\usepackage{graphics}
\usepackage{adjustbox}
\usepackage{csquotes}
\usepackage{mathrsfs,amsmath} 
\usepackage{nicefrac}
\usepackage{booktabs}
\usepackage[subtle]{savetrees}

\usepackage{pgfplotstable} 
\usepackage{graphicx}
\usepackage{pgfplots}
\usepgfplotslibrary{statistics}
\usepackage{graphics}
\usepackage{adjustbox}
\usetikzlibrary{pgfplots.groupplots}
\pgfplotsset{compat=1.11}
\usepgfplotslibrary{ternary}
\usepackage{tikz}
\usetikzlibrary{arrows.meta}
\usepackage{tkz-graph}
\usetikzlibrary{positioning,automata}
\usetikzlibrary{calc,spy,shapes}
\usetikzlibrary{arrows,decorations.pathmorphing,fit,positioning,patterns}
\pgfplotsset{compat=newest}

\emnlpfinalcopy

\def\emnlppaperid{***}

\DeclareFontFamily{U}{mathb}{\hyphenchar\font45}
\DeclareFontShape{U}{mathb}{m}{n}{
<-6> mathb5 <6-7> mathb6 <7-8> mathb7
<8-9> mathb8 <9-10> mathb9
<10-12> mathb10 <12-> mathb12
}{}
\DeclareSymbolFont{mathb}{U}{mathb}{m}{n}

\DeclareMathSymbol{\smalltriangleup} {2}{mathb}{"98}
\DeclareMathSymbol{\smalltriangledown} {2}{mathb}{"99}
\DeclareMathSymbol{\smalltriangleleft} {2}{mathb}{"9A}
\DeclareMathSymbol{\smalltriangleright}{2}{mathb}{"9B}
\DeclareMathSymbol{\blacktriangleup} {2}{mathb}{"9C}
\DeclareMathSymbol{\blacktriangledown} {2}{mathb}{"9D}
\DeclareMathSymbol{\blacktriangleleft} {2}{mathb}{"9E}
\DeclareMathSymbol{\blacktriangleright}{2}{mathb}{"9F}
\newcommand{\mypar}[1]{\medskip\noindent\textbf{#1}~}
\renewcommand{\footnotesize}{\scriptsize}

\usepackage{xspace} 
\newcommand{\tnet}{\textit{target network}\xspace}

\newcommand{\cnet}{\textit{confidence network}\xspace}
\newcommand{\wa}{\textit{weak annotator}\xspace}

\title{Avoiding Your Teacher's Mistakes:\\
Training Neural Networks with Controlled Weak Supervision}

\author{
Mostafa Dehghani\(^1\)
\(\quad\)
Aliaksei Severyn\(^2\)
\(\quad\)
Sascha Rothe\(^2\)
\(\quad\)
Jaap Kamps\(^1\)\\[1ex]
\mbox{}\(^1\)~University of Amsterdam
\\
\mbox{}\(^2\)~Google Research
\\
{\tt dehghani@uva.nl, severyn@google.com, rothe@google.com, kamps@uva.nl}}

\date{}

\begin{document}

\maketitle

\begin{abstract}
%

In this paper, we propose a semi-supervised learning method where we train two neural networks in a multi-task fashion: a \tnet and a \cnet. 
The \tnet is optimized to perform a given task and is trained using a large set of unlabeled data that are weakly annotated. 
We propose to weight the gradient updates to the \tnet using the scores provided by the second \cnet, which is trained on a small amount of supervised data. 
Thus we avoid that the weight updates computed from noisy labels harm the quality of the \tnet model.
We evaluate our learning strategy on two different tasks: document ranking and sentiment classification. The results demonstrate that our approach not only enhances the performance compared to the baselines but also speeds up the learning process from weak labels.
\end{abstract}

\section{Introduction}
\label{sec:introduction}
Deep neural networks have shown impressive results in a lot of tasks in computer vision, natural language processing, and information retrieval. However, their success is conditioned on the availability of exhaustive amounts of labeled data, while for many tasks such a data is not available. 
Hence, unsupervised and semi-supervised methods are becoming increasingly attractive. 

Using weak or noisy supervision is a straightforward approach to increase the size of the training data. 
For instance in web search, for the task of ranking, the ideal training data would be rankings of documents ordered by relevance for a large set of queries. However, it is not practical to collect such a data in large scale and only a small set of judged query-document pairs is available. However, for this task, the output of heuristic methods~\citep{Dehghani:2017:SIGIR} or clickthrough logs~\citep{Joachims:2002} can be used as weak or noisy signals along with a small amount of labeled data to train learning to rank models.

This is usually done by pre-training the network on weak data and fine-tuning it with true labels~\cite{Dehghani:2017:SIGIR, Severyn:2015:SIGIR}. 
However, these two independent stages do not leverage the full capacity of information from true labels. 
For instance, in the pet-raining stage there is no handle to control the extent to which the data with weak labels contribute in the learning process, while they can be of different quality.

In this paper, we propose a semi-supervised method that leverages a small amount of data with true labels along with a large amount of data with weak labels. 
Our proposed method has three main components:
A \wa, which can be a heuristic model, a weak classifier, or even human via crowdsourcing and it is employed to annotate massive amount of unlabeled data, a \tnet which uses a large set of weakly annotated instances by \wa to learn the main task, and  a \cnet which is trained on a small human-labeled set to estimate confidence scores for instances annotated by \wa. We train \tnet and \cnet in a multi-task fashion.

In a joint learning process, \tnet and \cnet try to learn a suitable representation of the data and this layer is shared between them as a two-way communication channel. 
The \tnet tries to learn to predict the label of the given input under the supervision of the \wa. In the same time, the output of \cnet, which are the confidence scores, define the magnitude of the weight updates to the \tnet with respect to the loss computed based on labels from \wa, during the back-propagation phase of the \tnet. This way, \cnet helps \tnet to avoid mistakes of her teacher, i.e. \wa, by down-weighting the weight updates from weak labels that do not look reliable to \cnet.

From a meta-learning perspective~\citep{dehghani2017:metalearning}, the goal of the \cnet trained jointly with the \tnet is to calibrate the learning rate for each instance in the batch. I.e., the weights $\pmb{w}$ of the \tnet $f_w$ at step $t+1$ are updated as follows:
\begin{equation}
\small
\pmb{w}_{t+1} = \pmb{w}_t - \frac{l_t}{b}\sum_{i=1}^b c_{\theta}(\tau_i, \tilde{y}_i)  \nabla \mathcal{L}(f_{\pmb{w_t}}(\tau_i), \tilde{y_i}) + \nabla \mathcal{R}(\pmb{w_t})
\end{equation}
where $l_t$ is the global learning rate, $b$ is the batch size, $\mathcal{L}(\cdot)$ the loss of predicting $\hat{y}=f_w(\tau)$ for an input $\pmb{\tau}$ when the target label is $\tilde{y}$; $c_\theta(\cdot)$ is a scoring function learned by the \cnet taking input instance $\pmb{\tau}_i$ and its noisy label $\tilde{y}_i$ and $\mathcal{R(.)}$ is the regularization term. 
Thus, we can effectively control the contribution to the parameter updates for the \tnet from weakly labeled instances based on how reliable their labels are according to the \cnet (learned on a small supervised data).

Our setup requires running a \wa to label a large amount of unlabeled data, which is done at pre-processing time. For many tasks, it is possible to use a simple heuristic, or implicit human feedback to generate weak labels. This set is then used to train the \tnet.  
In contrast, a small expert-labeled set is used to train the \cnet, which estimates how good the weak annotations are, i.e. controls the effect of weak labels on updating the parameters of the \tnet. 

Our method allows learning different types of neural architectures and different tasks, where a meaningful \wa is available. 
In this paper, we study the performance of our proposed model by focusing on two applications in information retrieval and natural language processing: document ranking and sentiment classification. 
Whilst these two applications differ considerably, as do the exact operationalization of our model to these cases, there are also some clear similarities.  
First, in both cases the human gold standard data is based on a cognitively complex, or subjective, judgments causing high interrater variation, increasing both the cost of obtaining labels as the need for larger sets of labels.
Second, also in both cases, the weak supervision signal is more systemic or objective, which facilitates the learning of the data representation.  

Our experimental results suggest that the proposed method is more effective in leveraging large amounts of weakly labeled data compared to traditional fine-tuning in both tasks. 
We also show that explicitly controlling the weight updates in the \tnet with the \cnet leads to faster convergence since the filtered supervision signals are more solid and less noisy.

In the following, in Section~\ref{sec:method}, we introduce the general architecture of our model and explain the training process. Then, we describe the details of the applications to which we apply our model in Section~\ref{sec:applications}. In Section~\ref{sec:exp} we present the experimental setups for each of the tasks along with its results and analysis. We then review related works and conclude the paper.

\section{The Proposed Method}
\label{sec:method}
In the following, we describe our recipe for semi-supervised learning of neural networks, in a scenario where along with a small human-labeled training set a large set of weakly labeled instances is leveraged.
Formally, given a set of unlabeled training instances, we run a \wa to generate weak labels.
This gives us the training set $U$. It consists of \emph{tuples} of training instances $\tau_i$ and their weak labels $\tilde{y}_i$, i.e. $U=\{(\tau_i, \tilde{y}_i),\ldots\}$. 

For a small set of training instances with true labels, we also apply the \wa to generate weak labels.
This creates the training set $V$, consisting of \emph{triplets} of training instances $\tau_j$, their weak labels $\tilde{y}_j$, and their true labels $y_j$, i.e. $V=\{(\tau_j,\tilde{y}_j,y_j),\ldots\}$. 
We can generate a large amount of training data $U$ at almost no cost using the \wa. 
In contrast, we have only a limited amount of data with true labels, i.e. $|V|<<|U|$. 

\begin{figure*}[!t]%
    \makebox[\textwidth][c]{
    \centering
    \begin{subfigure}[t]{0.5\textwidth}
        \centering
        \includegraphics[width=\textwidth]{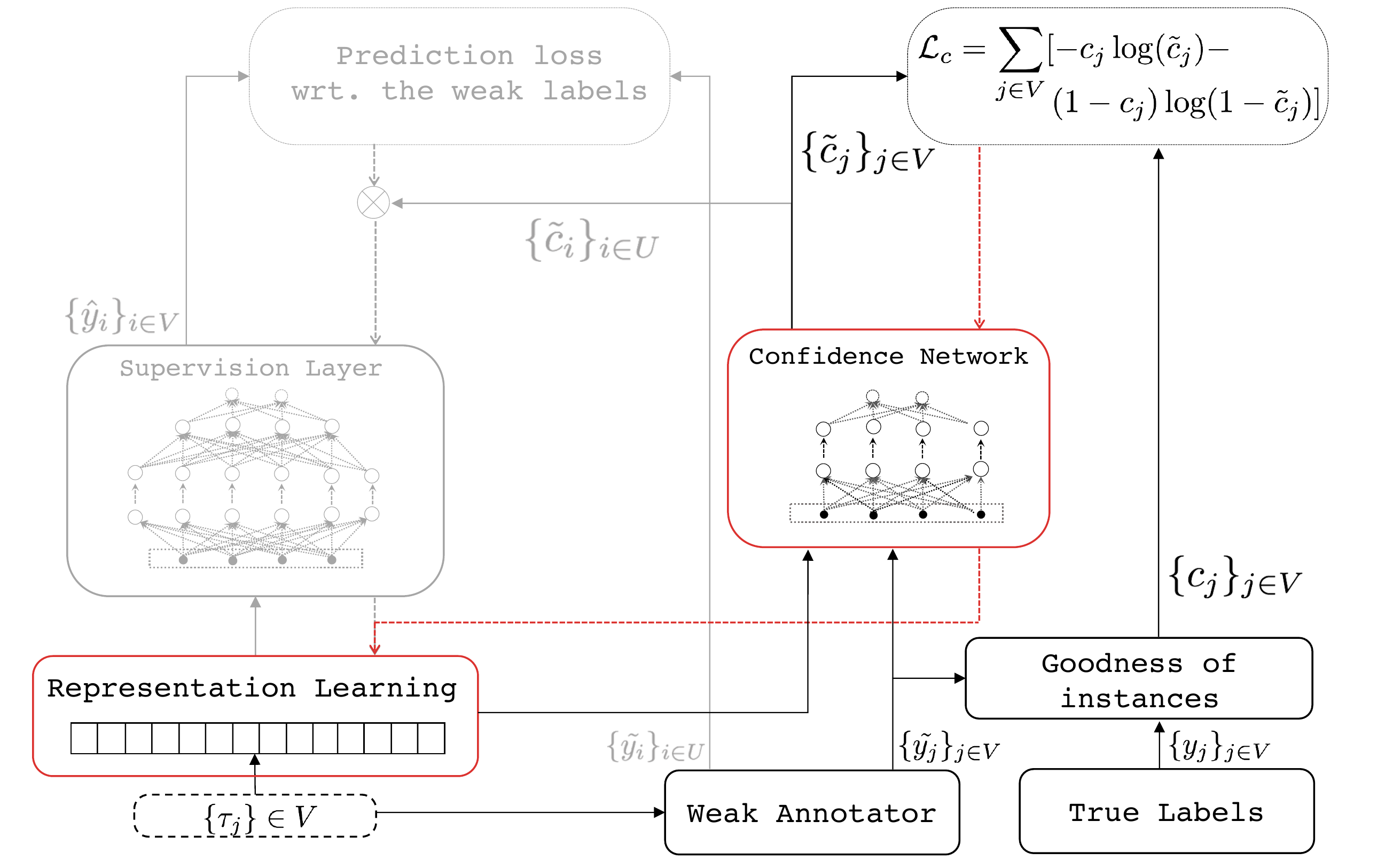}
        \vspace{-8pt}
        \caption{\label{fig:train_u}\footnotesize{Full Supervision Mode: Training on batches of data with true labels.}}
        \vspace{-4pt}
    \end{subfigure}%
    ~
    \begin{subfigure}[t]{0.5\textwidth}
        \centering
        \includegraphics[width=\textwidth]{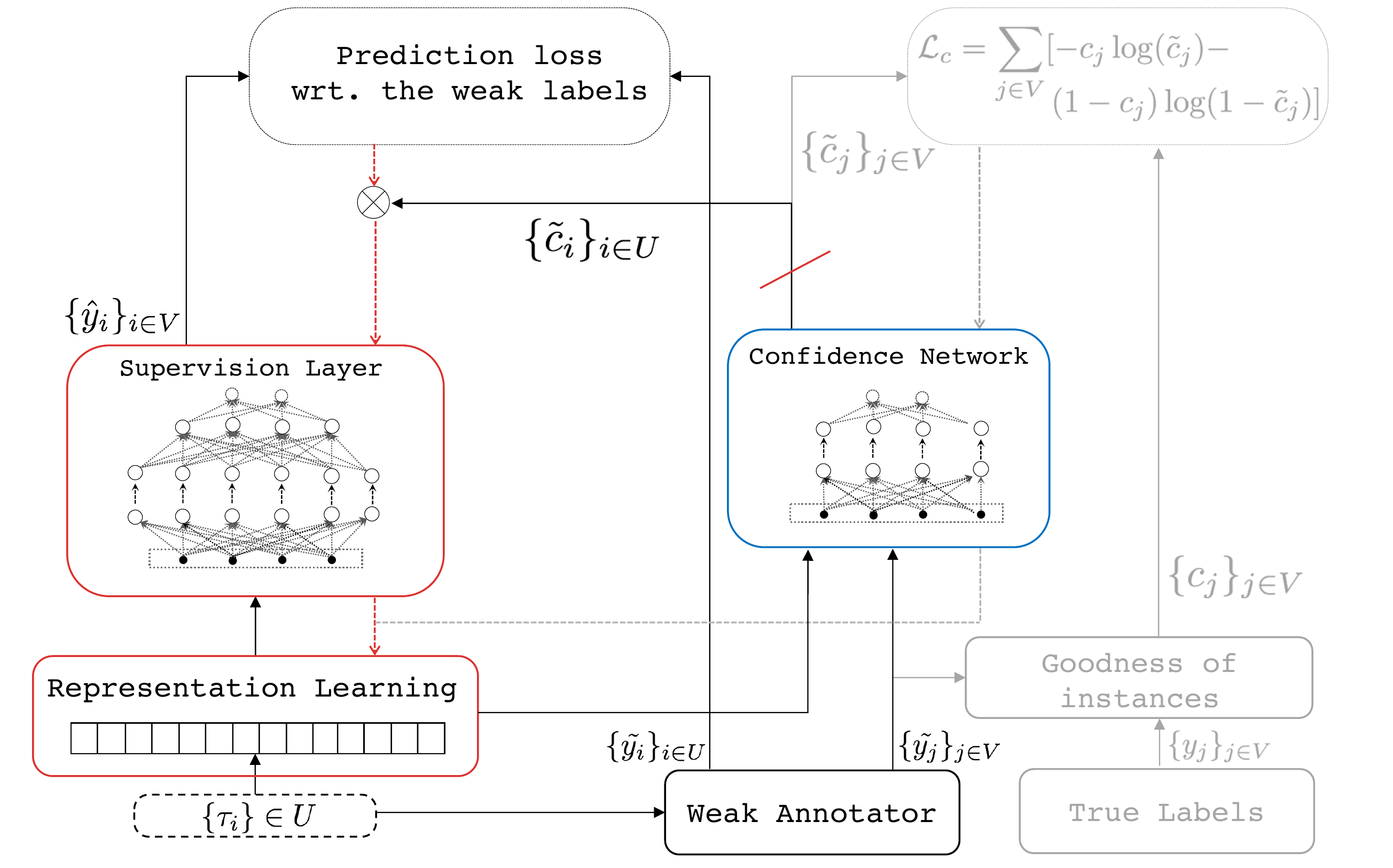}
        \vspace{-8pt}
        \caption{\label{fig:train_v}\footnotesize{Weak Supervision Mode: Training on batches of data with weak labels.}}
        \vspace{-4pt}
    \end{subfigure}%
    }
    \caption{\small{Learning from controlled weak supervision: Our proposed multi-task network for learning a target task in a semi-supervised fashion, using a large amount of weakly labeled data and a small amount of data with true labels.
    Faded parts of the network are disabled during the training in the corresponding mode. Red-dotted arrows show gradient propagation. Parameters of the parts of the network in red frames get updated in the backward pass, while parameters of the network in blue frames are fixed during the training.}}
    \label{fig:model}
    \vspace{-10pt}
\end{figure*}
\subsection{General Architecture}
\label{sec:generalarchitecture}
In our proposed framework we train a multi-task neural network that jointly learns the confidence score of weak training instances and the main task using controlled supervised signals.
The high-level representation of the model is shown in Figure~\ref{fig:model}: it comprises a \wa and two neural networks, namely the \cnet and the \tnet. 

The goal of the \wa is to \emph{provide weak labels} $\tilde{y}_i$ for all the instances $\tau_i \in U \cup V$. We have this assumption that $\tilde{y}_i$ provided by the \wa are imperfect estimates of true labels $y_i$, where $y_i$ are available for set $V$, but not for set $U$.

The goal of the \cnet is to \emph{estimate the confidence score} $\tilde{c}_j$ of training instances. It is learned on triplets from training set $V$: input $\tau_j$, its weak label $\tilde{y}_j$, and its true label $y_j$.
The score $\tilde{c}_j$ is then used to control the effect of weakly annotated training instances on updating the parameters of the \tnet in its backward pass during backpropagation.

The \tnet is in charge of \emph{handling the main task} we want to learn, or in other words, approximating the underlying function that predicts the correct labels. 
Given the data instance, $\tau_i$ and its weak label $\tilde{y}_i$ from the training set $U$, the \tnet aims to predict the label $\hat{y}_i$. 
The \tnet parameter updates are based on noisy labels assigned by the \wa, but the magnitude of the gradient update is based on the output of the \cnet. 

Both networks are trained in a multi-task fashion alternating between the \emph{full supervision} and the \emph{weak supervision} mode.  
In the \emph{full supervision} mode, the parameters of the \cnet get updated using batches of instances from training set $V$.  
As depicted in Figure~\ref{fig:train_v}, each training instance is passed through the representation layer mapping inputs to vectors. These vectors are concatenated with their corresponding weak labels $\tilde{y}_j$ generated by the \wa.
The \cnet then estimates $\tilde{c}_j$, which is the probability of taking data instance $j$ into account for training the \tnet.

In the \emph{weak supervision,} mode the parameters of the \tnet are updated using training set $U$.
As shown in Figure~\ref{fig:train_u}, each training instance is passed through the same representation learning layer and is then processed by the supervision layer which is a part of the \tnet predicting the label for the main task. 
We also pass the learned representation of each training instance along with its corresponding label generated by the \wa to the \cnet to estimate the \emph{confidence score} of the training instance, i.e. $\tilde{c}_i$. 
The confidence score is computed for each instance from set $U$. These confidence scores are used to weight the gradient updating \tnet parameters or in other words the step size during back-propagation. 

It is noteworthy that the representation layer is shared between both networks, so besides the regularization effect of layer sharing which leads to better generalization, sharing this layer lays the ground for the \cnet to benefit from the largeness of set $U$ and the \tnet to utilize the quality of set $V$. 

\subsection{Model Training}
\label{sec:modeltraining}
Our optimization objective is composed of two terms: (1) the \cnet loss $\mathcal{L}_c$, which captures the quality of the output from the \cnet and (2) the \tnet loss $\mathcal{L}_t$, which expresses the quality for the main task. 

Both networks are trained by alternating between the \emph{weak supervision} and the \emph{full supervision} mode.
In the \emph{full supervision} mode, the parameters of the \cnet are updated using training instance drawn from training set $V$. We use cross-entropy loss function for the \cnet to capture the difference between the predicted confidence score of instance $j$, i.e. $\tilde{c}_j$ and the target score $c_j$:
\begin{equation}
\mathcal{L}_c = \sum_{j\in V} -  c_j \log(\tilde{c}_j) - (1-c_j) \log(1-\tilde{c}_j),
\end{equation}
The target score $c_j$ is calculated based on the difference of the true and weak labels with respect to the main task.

In the \emph{weak supervision} mode, the parameters of the \tnet are updated using training instances from $U$. We use a weighted loss function, $\mathcal{L}_t$, to capture the difference between the predicted label $\hat{y}_i$ by the \tnet and target label $\tilde{y}_i$:
\begin{equation}
\mathcal{L}_t = \sum_{i\in U} \tilde{c}_i \mathcal{L}_i,
\end{equation}
where $\mathcal{L}_i$ is the task-specific loss on training instance $i$ and $\tilde{c}_i$ is the confidence score of the weakly annotated instance $i$, estimated by the \cnet.
Note that $\tilde{c}_i$ is treated as a constant during the weak supervision mode and there is no gradient propagation to the \cnet in the backward pass (as depicted in Figure~\ref{fig:train_u}). 

We minimize two loss functions jointly by randomly alternating between full and weak supervision modes (for example, using a 1:10 ratio).
During training and based on the chosen supervision mode, we sample a batch of training instances from $V$ with replacement or from $U$ without replacement (since we can generate as much train data for set $U$). Since in our setups usually $|U| >> |V|$, the training process oversamples the instance from $V$.

The key point here is that the ``main task'' and ``confidence scoring'' task are always defined to be close tasks and sharing representation will benefit the confidence network as an implicit data augmentation to compensate the small amount of data with true labels.
Besides, we noticed that updating the representation layer with respect to the loss of the other network acts as a regularization for each of these networks and helps generalization for both target and confidence network since we try to capture all tasks (which are related tasks) and less chance for overfitting.

We also investigated other possible setups or training scenarios. For instance, we tried updating the parameters of the supervision layer of the \tnet using also data with true labels. Or instead of using alternating sampling, we tried training the \tnet using controlled weak supervision signals after the \cnet is fully trained.
As shown in the experiments the architecture and training strategy described above provide the best performance.

\section{Applications}
\label{sec:applications}
In this section, we apply our semi-supervised method to two different tasks: \emph{document ranking} and \emph{sentiment classification}.
For each task, we start with an introduction of the task, followed by the setup of the target network, i.e. description of the representation learning layer and the supervision layer.

\subsection{Document Ranking}
This task is the core information retrieval problem which is challenging as it needs to capture the notion of relevance between query and documents. We employ a state-of-the-art pairwise neural ranker architecture as \tnet~\citep{Dehghani:2017:SIGIR}. 
In this setting, each training instance $\tau$ consists of a query $q$, and two documents $d^+$ and $d^-$.
The labels, $\tilde{y}$ and $y$, are scalar values indicating the probability of $d^+$ being ranked higher than $d^-$ with respect to $q$.

\begin{figure}[!t]%
    \centering
            \includegraphics[width=0.6\columnwidth]{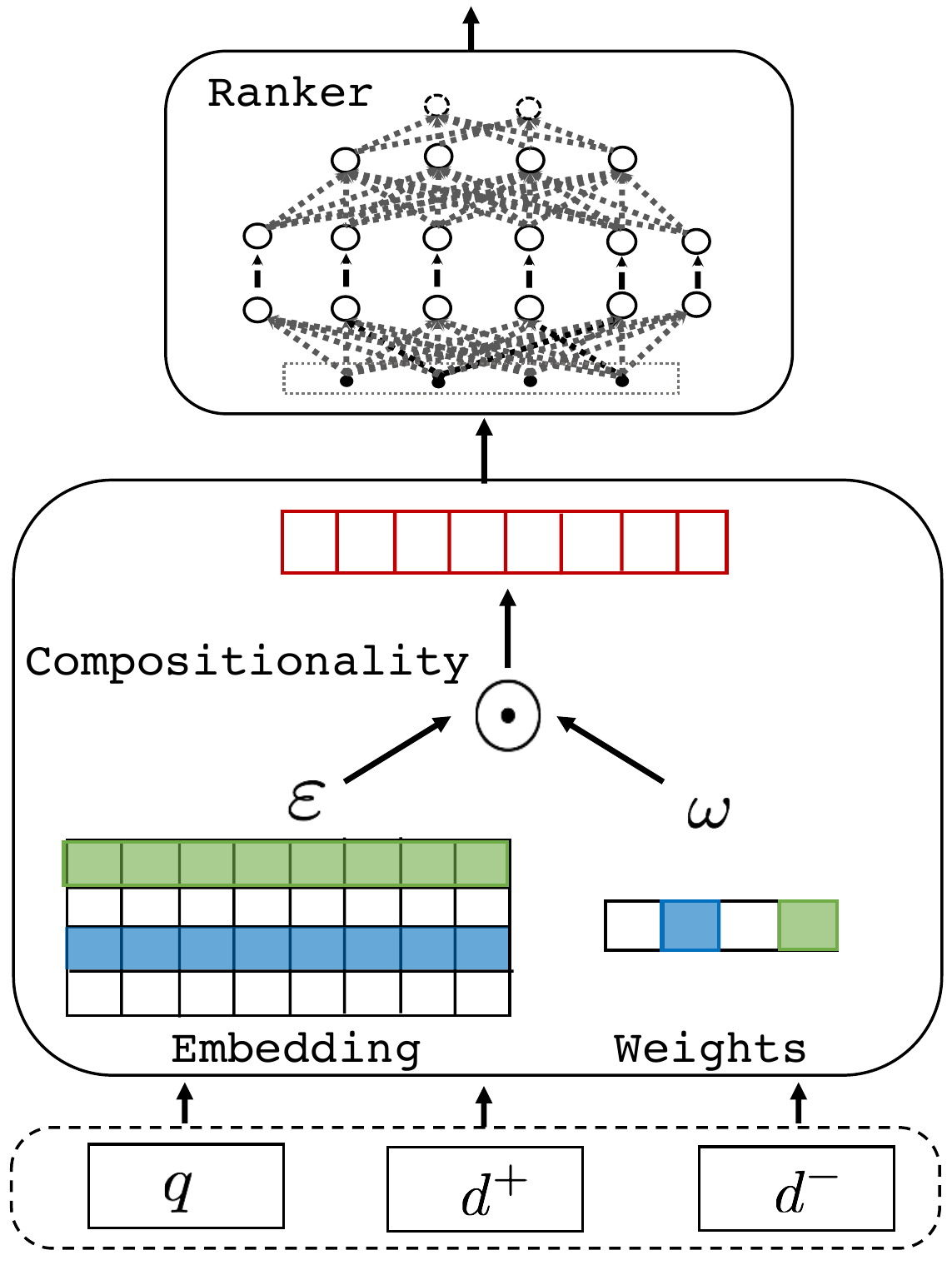}
     \vspace{-5pt}
    \caption{\small{The \tnet for the document ranking.}}
            \vspace{-10pt}
    \label{fig:ranker}
\end{figure}

The general schema of the \tnet is illustrated in Figure~\ref{fig:ranker}.

\mypar{The Representation Learning Layer} is a setup proposed in~\cite{Dehghani:2017:SIGIR}. This layer is a function $\psi$, which learns the representation of the input data instances, i.e. $(q, d^+, d^-)$, and consists of three components: (1) an embedding function $\varepsilon: \mathcal{V} \rightarrow \mathbb{R}^{m}$ (where $\mathcal{V}$ denotes the vocabulary set and $m$ is the number of embedding dimensions), (2) a weighting function $\omega: \mathcal{V} \rightarrow \mathbb{R}$, and (3) a compositionality function $\odot: (\mathbb{R}^{m}, \mathbb{R})^n \rightarrow \mathbb{R}^{m}$. More formally, the function $\psi$ is defined as:
\begin{equation}
     \begin{aligned}
\psi(q, d^+, d^-) = [ & \odot_{i=1}^{|q|}(\varepsilon(t_i^q),\omega(t_i^q)) ~ ||
& \\ &
\odot_{i=1}^{|d^+|} (\varepsilon(t_i^{d^+}),\omega(t_i^{d^+})) ~||
& \\ &
\odot_{i=1}^{|d^-|} (\varepsilon(t_i^{d^-}),\omega(t_i^{d^-})) ~],
     \end{aligned}     
\end{equation}
where $t_i^q$ and $t_i^d$ denote the $i^{th}$ term in query $q$ respectively document $d$.
The embedding function $\varepsilon$ maps each term to a dense $m$-\:dimensional real value vector, which is learned during the training phase.
The weighting function $\omega$ assigns a weight to each term in the vocabulary.

The compositionality function $\odot$ projects a set of $n$ embedding-weighting pairs to an $m$-\:dimensional representation, independent from the value of $n$:
\begin{equation}
\bigodot_{i=1}^n(\varepsilon(t_i),\omega(t_i)) = \frac{\sum_{i=1}^n\exp(\omega(t_i))\cdot \varepsilon(t_i)}{\sum_{j=1}^n{ \exp(\omega(t_j))}},
\end{equation}
which is in fact the normalized weighted element-wise summation of the terms' embedding vectors.

It has been shown that having global term weighting function along with embedding function improves the performance of ranking as it simulates the effect of inverse document frequency (IDF), which is an important feature in information retrieval~\cite{Dehghani:2017:SIGIR}.
In our experiments, we initialize the embedding function $\varepsilon$ with word2vec embeddings~\cite{Mikolov:2013} pre-trained on Google News and the weighting function $\omega$ with IDF.

\mypar{The Supervision Layer} receives the vector representation of the inputs processed by the representation learning layer and outputs a prediction $\tilde{y}$.
We opt for a simple fully connected feed-forward network with $l$ hidden layers followed by a softmax. 
Each hidden layer $z_k$ in this network computes $z_k = \alpha(W_k z_{k-1} + b_k)$, where $W_k$ and $b_k$ denote the weight matrix and the bias term corresponding to the $k^{th}$ hidden layer and $\alpha(.)$ is the non-linearity. These layers follow a sigmoid output. We employ the weighted cross entropy loss:
\begin{equation}
\mathcal{L}_t = \sum_{i\in B_U} \tilde{c}_i [- \tilde{y}_i \log (\hat{y}_i) - (1-\tilde{y}_i) \log(1-\hat{y}_i)],
\end{equation}
where $B_U$ is a batch of instances from $U$, and $\tilde{c}_i$ is the confidence score of the weakly annotated instance $i$, estimated by the \cnet.

\mypar{The Weak Annotator} is BM25~\cite{Robertson:2009} which is a well-performing unsupervised retrieval method. In the pairwise documents ranking setup, $\tilde{y}_i$ for a given instance $\tau_j = (q,d^+,d^-)$ is the probability of document $d^+$ being ranked higher than $d^-$, based on the scores obtained from the annotator: 
\begin{equation}
\tilde{y}_i = P_{q,d^+,d^-} = \frac{s_{q,d^+}}{s_{q,d^+} + s_{q,d^-}},
\end{equation}
whereas $s_{q,d}$ is the score obtained from the \wa.
To train the \cnet, the target label $c_j$ is calculated using the absolute difference of the true label and the weak label: $c_j= 1-|y_j - \tilde{y}_j|$, where $y_j$ is calculated similar to $\tilde{y}_i$, but $s_{q,d}$ comes from true labels created by humans.

\subsection{Sentiment Classification}
\begin{figure}[!t]%
    \centering
        \includegraphics[width=0.6\columnwidth]{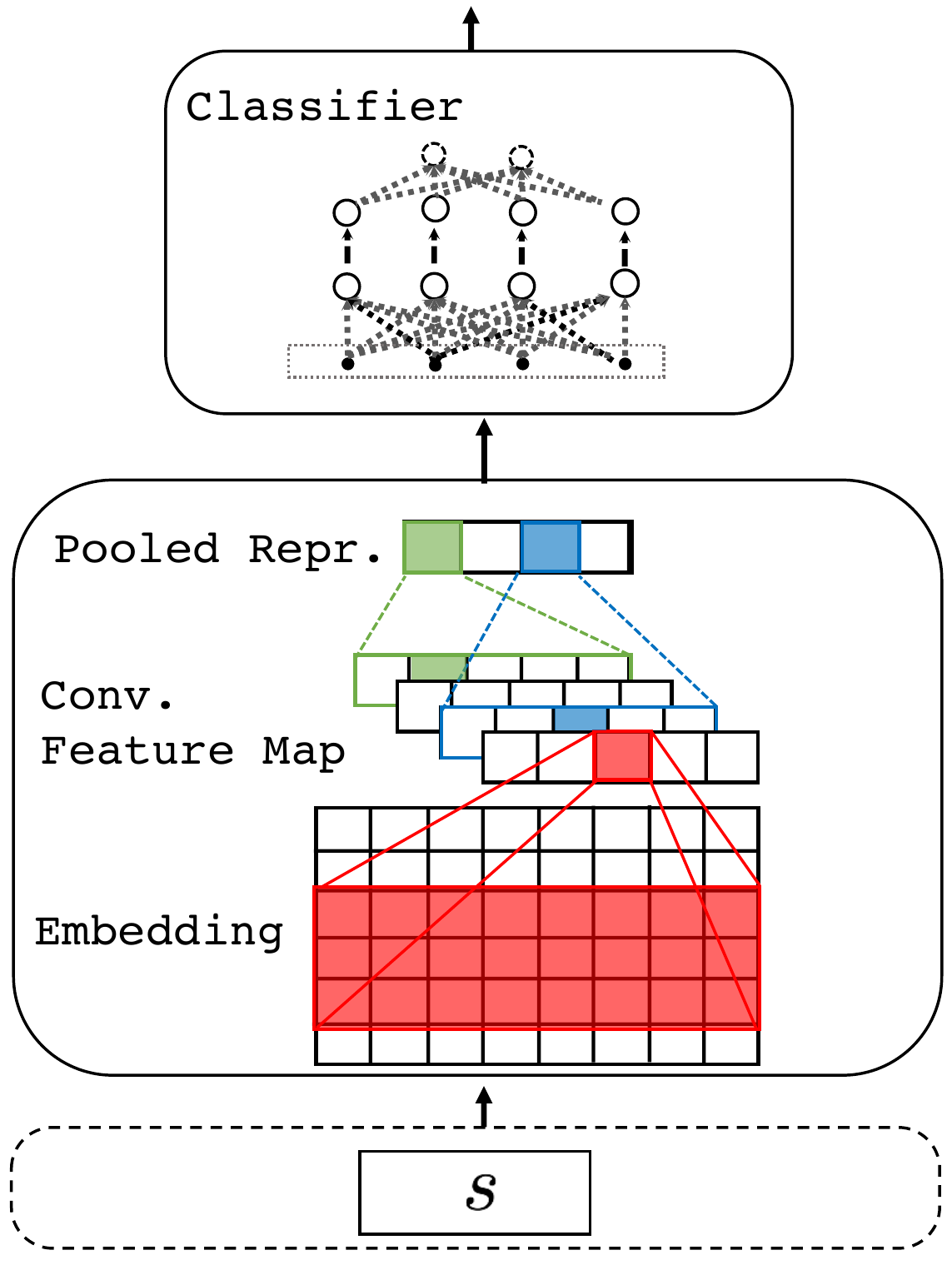}
        \vspace{-5pt}
    \caption{\small{The \tnet for the sentiment classification.}}
        \vspace{-10pt}
    \label{fig:sentiment}
\end{figure}

This task aims to identify the sentiment (e.g., positive, negative, or neutral) underlying an individual sentence.
Our \tnet is a convolutional model similar to~\citep{Deriu:2017, Severyn:2015:SIGIR, Severyn:2015:SemEval,Deriu2016:SemEval}. 
Each training instance $\tau$ consists of a sentence $s$ and its sentiment label $\tilde{y}$. The architecture of the \tnet is illustrated in Figure~\ref{fig:sentiment}

\mypar{The Representation Learning Layer} learns a representation for the input sentence $s$ and is shared between the \tnet and \cnet.
It consists of an embedding function $\varepsilon: \mathcal{V} \rightarrow \mathbb{R}^{m}$, where $\mathcal{V}$ denotes the vocabulary set and $m$ is the number of embedding dimensions.

This function maps the sentence to a matrix $S \in \mathbb{R}^{m \times |s|}$, where each column represents the embedding of a word at the corresponding position in the sentence. Matrix $S$ is passed through a convolution layer. 
In this layer, a set of $f$ filters is applied to a sliding window of length $h$ over $S$ to generate a feature map matrix $O$. Each feature map $o_i$ for a given filter $F$ is generated by $o_i = \sum_{k,j}S[i:i+h]_{k,j} F_{k,j}$, where $S[i:i+h]$ denotes the concatenation of word vectors from position $i$ to $i+h$. The concatenation of all $o_i$ produces a feature vector $o \in \mathbb{R}^{|s|-h+1}$. The vectors $o$ are then aggregated over all $f$ filters into a feature map matrix $O \in \mathbb{R}^{f\times(|s|-h+1)}$.

We also add a bias vector $b \in R^f$ to the result of a convolution.
Each convolutional layer is followed by a non-linear activation function (we use ReLU\cite{Nair:2010}) which is applied element-wise. Afterward, the output is passed to the max pooling layer which operates on columns of the feature map matrix $O$ returning the largest value: $pool(o_i) : \mathbb{R}^{1\times(|s|-h+1)} \rightarrow \mathbb{R}$ (see Figure~\ref{fig:sentiment}). 
This architecture is similar to the state-of-the-art model for Twitter sentiment classification from Semeval 2015 and 2016~\cite{Severyn:2015:SemEval,Deriu2016:SemEval}.

We initialize the embedding matrix with word2vec embeddings~\cite{Mikolov:2013} pretrained on a collection of 50M tweets.

\mypar{The Supervision Layer} is a feed-forward neural network similar to the supervision layer in the ranking task (with different width and depth) but with softmax instead of sigmoid as the output layer which returns $\hat{y}_i$, the probability distribution over all three classes. 
We employ the weighted cross entropy loss:
\begin{equation}
\mathcal{L}_t = \sum_{i\in B_U} \tilde{c}_i \sum_{k \in K} - \tilde{y}_i^k \log (\hat{y}_i^k),
\end{equation}
where $B_U$ is a batch of instances from $U$, and $\tilde{c}_i$ is the confidence score of the weakly annotated instance $i$, and $K$ is a set of classes.

\mypar{The Weak Annotator}\label{sentiment-WA} for the sentiment classification task is a simple unsupervised lexicon-based method~\cite{Hamdan:2013,Kiritchenko:2014}.
We use SentiWordNet03~\cite{Gaccianella:2010} to assign probabilities (positive, negative and neutral) for each token in set $U$.
Then a sentence-level distribution is derived by simply averaging the distributions of the terms, yielding a noisy label $\tilde{y}_i \in \mathbb{R}^{|K|}$, where $|K|$ is the number of classes, i.e. $|K|=3$. 
We empirically found that using soft labels from the \wa works better than assigning a single hard label.
The target label $c_j$ for the \cnet is calculated by using the mean absolute difference of the true label and the weak label: $c_j= 1-\frac{1}{|K|}\sum_{k\in K}|y_j^k - \tilde{y}_j^k|$, where $y_j$ is the one-hot encoding of the sentence label over all classes.

\section{Experiments and Results}
\label{sec:exp}
Here we first describe baselines. Afterward, we present the experimental setups for each of our tasks along with their results and analysis.

\subsection{Baselines and General Setups}
For both tasks, we evaluate the performance of our method compared to the following baselines:
\begin{itemize}[leftmargin=*]
\setlength{\topsep}{0.1pt}
\setlength{\partopsep}{0.1pt}
\setlength{\itemsep}{0.1pt}
\setlength{\parskip}{0.1pt}
\setlength{\parsep}{0.1pt}
\item 
(1.WA) Weak Annotator, i.e. the unsupervised method that we used for annotating the unlabeled data.
\item
(2.WSO) Weak Supervision Only, i.e. the \tnet trained only on weakly labeled data.
\item
(3.FSO) Full Supervision Only, i.e. the \tnet trained only on true labeled data.
\item
(4.WS+FT) Weak Supervision + Fine Tuning, i.e. the \tnet trained on the weakly labeled data and fine-tuned on true labeled data.
\item
(5.WS+SFT) Weak Supervision + Supervision Layer Fine-Tuning, i.e. the \tnet trained only on weakly labeled data and the supervision layer is fine-tuned on true labeled data while the representation learning layer is kept fixed.
\item
(6.WS+RFT) Weak Supervision + Representation Fine Tuning, i.e. WS+SFT, except the supervision layer is kept fixed during fine tuning.
\item
(7.NLI) New Label Inference~\cite{Veit:2017} is similar to our proposed neural architecture inspired by the teacher-student paradigm~\cite{Hinton:2015,Romero:2014}, but instead of having the \cnet to predict the ``confidence score'' of the training instance, there is a \emph{label generator network} which is trained on set $V$ to map the weak labels of the instances in $U$ to the \emph{new labels}. The new labels are then used as the target for training the \tnet.
\item
(8.CWS$_\text{JT}$) Controlled Weak Supervision with Joint Training is our proposed neural architecture in which we jointly train the \tnet and the \cnet by alternating batches drawn from sets $V$ and $U$ (as explained in Section~\ref{sec:modeltraining}). 
\item
(9.CWS$_\text{JT+}$) Controlled Weak Supervision + Full Supervision with Joint Training is the same as CWS$_\text{JT}$, except that parameters of the supervision layer in \tnet are also updated using batches from $V$, with regards to the true labels.
\end{itemize}

Additionally, we compare the performance of CWS$_\text{JT}$, with other possible training setups:
\begin{itemize}[leftmargin=*]
\setlength{\topsep}{0.1pt}
\setlength{\partopsep}{0.1pt}
\setlength{\itemsep}{0.1pt}
\setlength{\parskip}{0.1pt}
\setlength{\parsep}{0.1pt}
\item 
(a.CWS$_\text{ST}$) Separate Training, i.e. we consider the \cnet as a separate network, without sharing the representation learning layer, and train it on set $V$. We then train the \tnet on the controlled weak supervision signals.
\item 
(b.CWS$_\text{CT}$) Circular Training, i.e. we train the \tnet on set $U$. Then the \cnet is trained on data with true labels, and the \tnet is trained again but on controlled weak supervision signals.
\item
(c.CWS$_\text{PT}$) Progressive Training is the mixture of the two previous baselines. Inspired by \cite{Rusu:2016}, we transfer the learned information from the converged \tnet to the \cnet using progressive training. We then train the \tnet again on the controlled weak supervision signals.
\end{itemize}

The proposed architectures are implemented in TensorFlow~\citep{tang2016:tflearn,tensorflow2015-whitepaper}. We use the Adam optimizer~\citep{Kingma:2014} and the back-propagation algorithm. Furthermore, to prevent feature co-adaptation, we use \emph{dropout}~\citep{Srivastava:2014} as a regularization technique in all models. 

In our setup, the \cnet to predict $\tilde{c}_j$ is a fully connected feed forward network.
Given that the \cnet is learned only from a small set of true labels and to speed up training we initialize the representation learning layer with pre-trained parameters, i.e., pre-trained word embeddings.
We use ReLU~\cite{Nair:2010} as a non-linear activation function $\alpha$ in both \tnet and \cnet.
In the following, we describe task-specific setups and the experimental results.

\subsection{Document Ranking Setup \& Results}
\mypar{Collections.}
We use two standard TREC collections for the task of ad-hoc retrieval: The first collection (\emph{Robust04}) consists of 500k news articles from different news agencies as a homogeneous collection. The second collection (\emph{ClueWeb}) is ClueWeb09 Category B, a large-scale web collection with over 50 million English documents, which is considered as a heterogeneous collection. Spam documents were filtered out using the Waterloo spam scorer~\footnote{\url{http://plg.uwaterloo.ca/~gvcormac/clueweb09spam/}}~\citep{Cormack:2011} with the default threshold $70\%$. 

\mypar{Data with true labels.} 
We take query sets that contain human-labeled judgments: a set of 250 queries (TREC topics 301--450 and 601--700) for the Robust04 collection and a set of 200 queries (topics 1-200) for the experiments on the ClueWeb collection.
For each query, we take all documents judged as relevant plus the same number of documents judged as non-relevant and form pairwise combinations among them.

\mypar{Data with weak labels.}
We create a query set $Q$ using the unique queries appearing in the AOL query logs~\citep{Pass:2006}.
This query set contains web queries initiated by real users in the AOL search engine that were sampled from a three-month period from March 2006 to May 2006. 
We applied standard pre-processing~\cite{Dehghani:2017:SIGIR,Dehghani2017:CIKM} on the queries. We filtered out a large volume of navigational queries containing URL substrings (``http'', ``www.'', ``.com'', ``.net'', ``.org'', ``.edu''). We also removed all non-alphanumeric characters from the queries. For each dataset, we took queries that have at least ten hits in the target corpus using our \wa method. Applying all these steps, 
We collect 6.15 million queries to train on in Robust04 and 6.87 million queries for ClueWeb.
To prepare the weakly labeled training set $U$, we take the top $1000$ retrieved documents using BM25 for each query from training query set $Q$, which in total leads to $\sim|Q|\times 10^6$ training instances. 

\mypar{Parameters and Settings.}
We conducted a nested 3-fold cross validation with $80/20$ training/validation split in each fold. All hyper-parameters of all models and baselines were tuned individually on the validation set using batched GP bandits with an expected improvement acquisition function~\citep{Desautels:2014}.
The size and number of hidden layers for the ranker and the \cnet were separately selected from $\{64, 128, 256, 512\}$ and $\{1, 2, 3, 4\}$, respectively. The initial learning rate and the dropout parameter were selected from $\{10^{-3}, 10^{-5}\}$ and $\{0.0, 0.2, 0.5\}$, respectively. We considered embedding sizes of $\{300, 500\}$. The batch size in our experiments was set to $128$.

In all experiments, the parameters of the network are optimized employing the Adam optimizer~\citep{Kingma:2014} and using the computed gradient of the loss to perform the back-propagation algorithm.
At inference time, for each query, we take the top $2000$ retrieved documents using BM25 as candidate documents and re-rank them using the trained models. We use the Indri\footnote{\url{https://www.lemurproject.org/indri.php}} implementation of BM25 with default parameters (i.e., $k_1 = 1.2$, $b = 0.75$, and $k_3 = 1000$).

\newcommand{\ps}{$^\blacktriangleup$}
\newcommand{\ns}{$^\smalltriangledown$}
\newcommand{\fs}{$^{~}$}

\begin{table}[tbp]
\caption{\label{tbl_main} Performance of the proposed method and baseline models on different datasets. 
\small{\normalfont{(\ps or \ns  indicates that the improvements or degradations are statistically significant, at the 0.05 level using the paired two-tailed t-test. For all model, the improvement/degradations is with respect to the ``weak supervision only'' baseline ({\small{WSO}}). For CWS$_\text{JT}$, the improvement over all baselines is considered and the Bonferroni correction is applied on the significant tests.)}}}
\vspace{-5pt}
\centering
\begin{adjustbox}{max width=\columnwidth}
\begin{tabular}{r l c c c c}
\toprule
& \multirow{2}{*}{Method} &
\multicolumn{2}{c}{Robust04} & \multicolumn{2}{c}{ClueWeb}
\\ \cmidrule(lr){3-4} \cmidrule(lr){5-6}
& & \small{MAP}\fs & \small{nDCG@20}
& \small{MAP}\fs & \small{nDCG@20}
\\ \midrule
1 & \small{WA$_\text{BM25}$} 
& 0.2503\fs & 0.4102\fs  
& 0.1021\fs & 0.2070\fs
\\ \midrule
2 & \small{WSO} 
& 0.2702\fs & 0.4290\fs  
& 0.1297\fs & 0.2201\fs
\\
3 & \small{FSO} 
& 0.1790\ns & 0.3519\ns  
& 0.0782\ns & 0.1730\ns
\\ \midrule
4 & \small{WS+FT} 
&  0.2830\ps & 0.4355\ps 
&  0.1346\ps & 0.2346\ps
\\
5 & \small{WS+SFT} 
&  0.2711\fs & 0.4203\fs 
&  0.1002\ns & 0.1940\ns
\\
6 & \small{WS+RFT} 
&  0.2810\ps & 0.4316\fs 
&  0.1286\fs & 0.2240\fs
\\ \midrule
7 & \small{NLI}
&  0.2421\ns & 0.4092\ns 
&  0.1010\ns & 0.2004\ns
\\
8 & \small{CWS$_\text{JT}$} 
& \textbf{0.3024}\ps  & \textbf{0.4507}\ps  
& \textbf{0.1372}\ps  & \textbf{0.2453}\ps 
\\
9 & \small{CWS$_\text{JT}^+$} 
& 0.2786\ps  & 0.4367\ps  
& 0.1310\fs  & 0.2244\fs 

\\\bottomrule
\end{tabular}
\vspace{-5pt}
\end{adjustbox}
\vspace{-10pt}
\end{table}

\mypar{Results and Discussions.}
We evaluate on set $V$ and report two standard evaluation metrics: mean average precision (MAP) of the top-ranked $1000$ documents and normalized discounted cumulative gain calculated for the top $20$ retrieved documents (nDCG@20).
Statistical significant differences of MAP and nDCG@20 values are determined using the two-tailed paired t-test with $p\_value<0.05$, with Bonferroni correction.

Table~\ref{tbl_main} shows the performance on both datasets. Based on the results, {\small ${CWS}_{JT}$} provides a significant boost on the performance over all datasets. 

There are two interesting points we want to highlight.
First, among the fine-tuning experiments, updating all parameters of the \tnet is the best fine tuning strategy. Updating only the parameters of the representation layer based on the true labels works better than updating only parameters of the supervision layer. This supports our designed choice of a shared embedding layer which gets updated on set $V$. 

Second, while it seems reasonable to make use of true labels for updating \emph{all} parameters of the \tnet, CWS$_\text{JT}^+$ achieves no better results than CWS$_\text{JT}$. It also performs mostly even worse than WS+FT. 
This is because during training, the direction of the parameter optimization is highly affected by the type of supervision signal and while we control the magnitude of the gradients, we do not change their directions, so alternating between two sets with different label qualities (different supervision signal types, i.e. weak and string) confuses the supervision layer of the \tnet. In fine tinning, we don not have this problem since we optimize the parameters with respect to the supervision from these two sets in two separate stages.

It is noteworthy that we have also tried CWS$_\text{JT}^+$ with another objective function for the \tnet taking both weak and true labels into account which was slightly better, but gives no improvement over CWS$_\text{JT}$.

In the ranking task, the \tnet is designed in particular to be trained on weak annotations~\cite{Dehghani:2017:SIGIR}, hence training the network only on weak supervision performs better than FSO. This is due to the fact that ranking is a complex task requiring many training instances, while relatively few true labels are available.

The performance of NLI is worse than CWS$_\text{JT}$ as learning a mapping from imperfect labels to accurate labels and training the \tnet on new labels is essentially harder than learning to filter out the noisy labels, hence needs a lot of supervised data. The reason is that for the ranking, due to a few training instances with regards to the task complexity, NLI fails to generate better new labels, hence it directly misleads the \tnet and completely fails to improve the performance. 

\begin{table}[tbp]
\centering
\caption{\label{tbl_variants}Performance of the variants of the proposed method on different datasets. 
\small{\normalfont{(\ps or\ns  indicates that the improvements or degradations are statistically significant, at the 0.05 level using the paired two-tailed t-test. For all model, the improvement/degradations is with respect to the ``weak supervision only'' baseline ({\small{WSO}} on Table~\ref{tbl_main}) . For CWS$_\text{JT}$, the improvement over all baselines is considered and the Bonferroni correction is applied on the significant tests.)}}}
\vspace{-5pt}
\begin{adjustbox}{max width=\columnwidth}
\begin{tabular}{r l c c c c}
\toprule
& \multirow{2}{*}{Method} &
\multicolumn{2}{c}{Robust04} & \multicolumn{2}{c}{ClueWeb}
\\ \cmidrule(lr){3-4} \cmidrule(lr){5-6}
& & \small{MAP} & \small{nDCG@20}
& \small{MAP} & \small{nDCG@20}
\\ \midrule
a & \small{CWS$_\text{ST}$} 
&  0.2716\fs  & 0.4237\fs 
&  0.1320\fs  & 0.2213\fs
\\ 
b & \small{CWS$_\text{CT}$} 
&  0.2961\ps & 0.4440\ps 
&  \textbf{0.1378}\ps  & 0.2431\ps 
\\ 
c & \small{CWS$_\text{PT}$} 
& 0.2784\ps  & 0.4292\fs  
& 0.1314\fs  & 0.2207\fs
\\ 
& \small{CWS$_\text{JT}$} 
&  \textbf{0.3024}\ps  & \textbf{0.4507}\ps  
&  0.1372\ps  & \textbf{0.2453}\ps 
\\ \bottomrule
\end{tabular}
\vspace{-5pt}
\end{adjustbox}
\vspace{-10pt}
\end{table}

Table~\ref{tbl_variants} shows the performance of different training strategies. 
As shown, CWS$_\text{JT}$ and CWS$_\text{CT}$ perform better than other strategies. CWS$_\text{CT}$ is to let the \cnet to be trained separately, while still being able to enjoy shared learned information from the \tnet. However, it is less efficient as we need two rounds of training on weakly labeled data.

CWS$_\text{ST}$ performs poorly since the training data $V$ is too small to train a high-quality \cnet without taking advantage of the vast amount of weakly annotated data in $U$. We also noticed that this strategy leads to a slow convergence compared to WSO.
Also transferring learned information from \tnet to \cnet via progressive training, i.e. CWS$_\text{PT}$, performs no better than full sharing of the representation learning layer.

\subsection{Sentiment Classification Setup \& Results}
\mypar{Collections.}
We test our model on the twitter message-level sentiment classification of SemEval-15 Task 10B \cite{rosenthal:2015}. Datasets of SemEval-15 subsume the test sets from previous editions of SemEval, i.e. SemEval-13 and SemEval-14. Each tweet was preprocessed so that URLs and usernames are masked.

\mypar{Data with true labels.} 
We use train (9,728 tweets) and development (1,654 tweets) data from SemEval-13 for training and SemEval-13-test (3,813 tweets) for validation.
To make our results comparable to the official runs on SemEval we use SemEval-14 (1,853 tweets) and  SemEval-15 (2,390 tweets) as test sets~\cite{rosenthal:2015, Nakov:2016}.

\mypar{Data with weak labels.}
We use a large corpus containing 50M tweets collected during two months for both, training the word embeddings and creating the weakly annotated set $U$ using the lexicon-based method explained in Section~\ref{sentiment-WA}. 

\begin{table}[tbp]
\centering
\caption{\label{tbl_main_sent}Performance of the baseline models as well as the proposed method on different datasets. \small{\normalfont{(\ps or\ns  indicates that the improvements or degradations are statistically significant, at the 0.05 level using the paired two-tailed t-test. For all model, the improvement/degradations is with respect to the ``weak supervision only'' baseline ({\small{WSO}}). For CWS$_\text{JT}$, the improvement over all baselines is considered and the Bonferroni correction is applied on the significant tests.)}}}
\vspace{-5pt}
\begin{adjustbox}{max width=0.75\columnwidth}
\begin{tabular}{r l c c}
\toprule
& Method & SemEval-14 & SemEval-15
\\ \midrule
1 & \small{WA$_\text{Lexicon}$} & 0.5141 & 0.4471
\\ \midrule
2 & \small{WSO} & 0.6719\fs & 0.5606\fs 
\\
3 & \small{FSO} & 0.6307\fs & 0.5811\fs
\\ \midrule
4 & \small{WS+FT} & 0.7080\ps & 0.6441\ps
\\
5 & \small{WS+SFT} & 0.6875\fs & 0.6193\ps
\\
6 & \small{WS+RFT} & 0.6932\fs & 0.6102\ps
\\ \midrule
7 & \small{NLI} & 0.7113\ps & 0.6433\ps
\\
8 & \small{CWS$_\text{JT}$} & \textbf{0.7362}\ps & \textbf{0.6626}\ps
\\
9 & \small{CWS$_\text{JT}^+$} & 0.7310\ps & 0.6551\ps
\\
& \small{SemEval$^\text{1th}$} & 0.7162\ps & 0.6618\ps
\\\bottomrule
\end{tabular}
\vspace{-5pt}
\end{adjustbox}
\vspace{-10pt}
\end{table}
\begin{table}[tbp]
\centering
\caption{\label{tbl_variants_sent} Performance of the variants of the proposed method for sentiment classification task, on different datasets.
\small{\normalfont{(\ps or\ns  indicates that the improvements or degradations are statistically significant, at the 0.05 level using the paired two-tailed t-test. For all model, the improvement/degradations is with respect to the ``weak supervision only'' baseline ({\small{WSO}} on Table~\ref{tbl_main_sent}) . For CWS$_\text{JT}$, the improvement over all baselines is considered and the Bonferroni correction is applied on the significant tests.)}}}
\vspace{-5pt}
\begin{adjustbox}{max width=0.75\columnwidth}
\begin{tabular}{r l c c}
\toprule
& Method & SemEval-14 & SemEval-15
\\ \midrule
a & \small{CWS$_\text{ST}$} & 0.7183\ps & 0.6501\ps
\\ 
b & \small{CWS$_\text{CT}$} & \textbf{0.7363}\ps & \textbf{0.6667}\ps
\\ 
c & \small{CWS$_\text{PT}$} & 0.7009\ps & 0.6118\ps
\\ 
& \small{CWS$_\text{JT}$} & 0.7362\ps & 0.6626\ps
\\\bottomrule
\end{tabular}
\vspace{-5pt}
\end{adjustbox}
\vspace{-10pt}
\end{table}

\mypar{Parameters and Settings.}
Similar to the document ranking task, we tuned the hyper-parameters for each model, including baselines, separately with respect to the true labels of the validation set using batched GP bandits with an expected improvement acquisition function~\citep{Desautels:2014}.  
The size and number of hidden layers for the classifier and the \cnet were separately selected from $\{32, 64, 128\}$ and $\{1, 2, 3\}$, respectively.
We tested the model with both, $1$ and $2$ convolutional layers. The number of convolutional feature maps and the filter width is selected from $\{200,300\}$ and $\{ 3, 4, 5\}$, respectively. The initial learning rate and the dropout parameter were selected from $\{1E-3, 1E-5\}$ and $\{0.0, 0.2, 0.5\}$, respectively. We considered embedding sizes of $\{100, 200\}$ and the batch size in these experiments was set to $64$.

\mypar{Results and Discussion.}
We report the performance of our model and the baseline models in terms of official SemEval metric, Macro-F1, in Table~\ref{tbl_main_sent}. 
We have also report statistical significance of F1 improvements using two-tailed paired t-test with $p\_value<0.05$, with Bonferroni correction. 
Our method is the best performing among all the baselines. 

Unlike the ranking task, training the network only on data with true labels, i.e. TSO, performs rather good. In the sentiment classification task, learning representation of input which is a sentence (tweet) is simpler than the ranking task in which we try to learn representation for query and long documents. Consequently, we need fewer data to be able to learn a suitable representation and with the amount of available data with true labels, we can already capture a rather good representation without helps of weak data, while it was impossible in the ranking task. 

However, as the results suggest, we can still gain improvement using fine-tuning. In this task, behaviors of different fine-tuning experiments are similar to the ranking task. Furthermore, updating parameters of the supervision layer, with respect to the true labels, i.e. CWS$_\text{JT}^+$ model, does not perform better than CWS$_\text{JT}$, which again supports our choice of updating just the representation learning layer with respect to the signals from data with true labels. 

In the sentiment classification task, the performance of NLI is acceptable compared to the ranking task. This is first of all because generating new classification labels is essentially simpler. Secondly, in this task, we need to learn to represent a simpler input, and learn a simpler function to predict the labels, but a relatively bigger set of supervised data which helps to generate new labels. 
However, the performance of NLI is still lower than CWS$_\text{JT}$. 
We can argue that CWS$_\text{JT}$ is a more conservative approach. It is in fact equipped with a \emph{soft} filter that decreases the effect of noisy training examples from set $U$ on parameter updates during training. 
This is a smoother action as we just down-weight the gradient, while NLI might change the direction of the gradient by generating a completely new label and consequently it is prone to more errors, especially when there is not enough high-quality training data to learn to generate better labels.

In the sentiment classification task, besides the general baselines, we also report the best performing systems, which are also convolution-based models (\citet{Rouvier:2016} on SemEval-14; \citet{Deriu2016:SemEval} on SemEval-15). Our proposed model outperforms the best system on both datasets.

Table~\ref{tbl_variants_sent} also presents the results of different training strategies for the sentiment classification task. 
As shown, similar to the ranking task, CWS$_\text{JT}$ and CWS$_\text{CT}$ perform better than other strategies. Although CWS$_\text{CT}$ is slightly better (not statistically significant) in terms of effectiveness compared to CWS$_\text{JT}$, it is not as efficient as CWS$_\text{JT}$ during training.

Compared to the ranking task, for sentiment classification, it is easier to estimate the confidence score of instances with respect to the amount of available supervised data. Therefore, CWS$_\text{ST}$ is able to improve the performance over WSO significantly. 
Moreover, CWS$_\text{PT}$ fails compared to the strategies where the representation learning layer is shared between the \tnet and the \cnet.

\begin{figure}[!t]%
    \centering
    \includegraphics[width=\columnwidth]{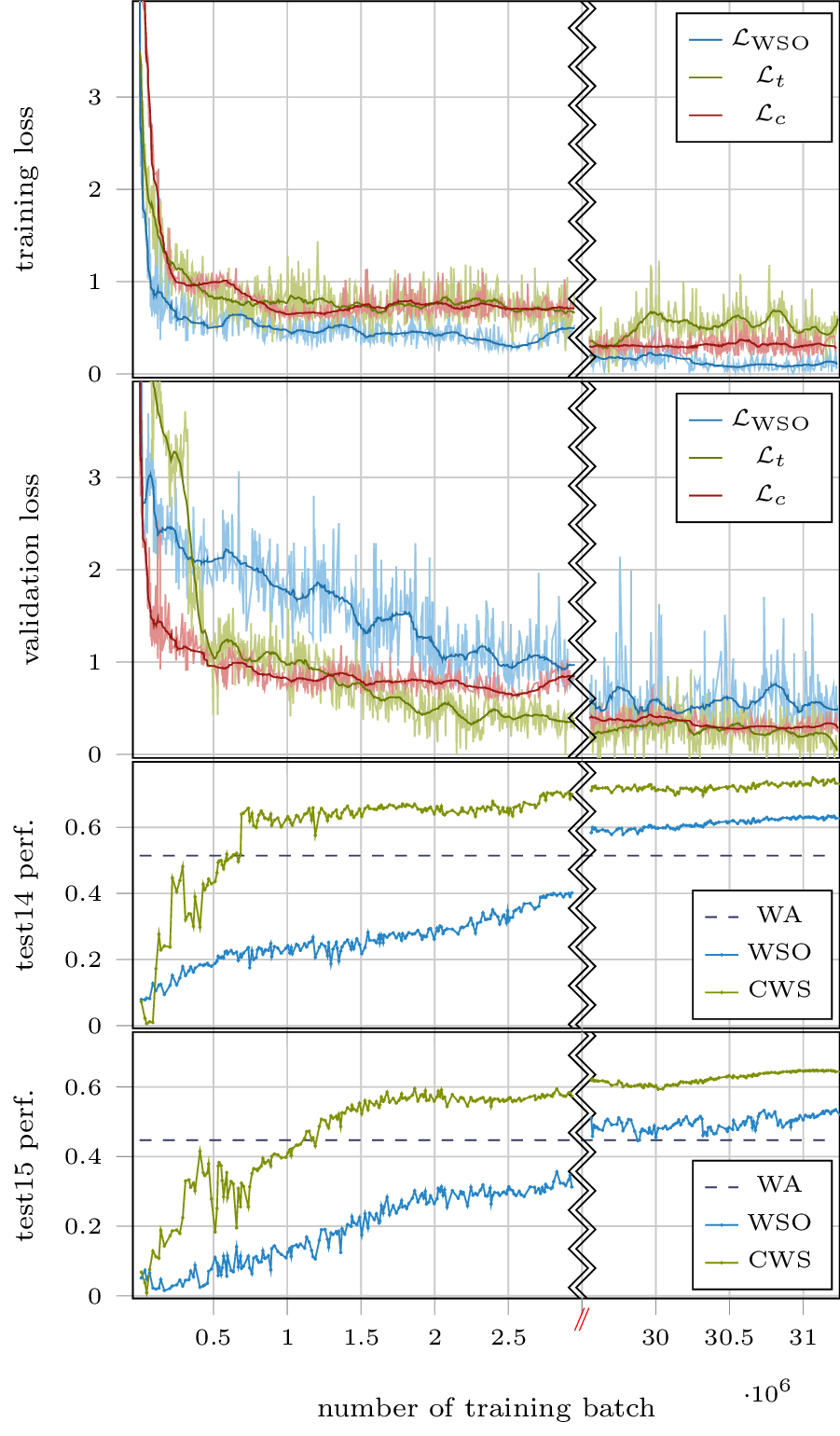}
    \vspace{-15pt}
    \caption{\small{Loss of the \tnet ($\mathcal{L}_t$) and the \cnet ($\mathcal{L}_c$) compared to the loss of WSO ($\mathcal{L}_\text{WSO}$) on training/validation set and performance of CWS, WSO, and WA on test sets with respect to different amount of training data on sentiment classification.}}
    \label{fig:plot}
    \vspace{-15pt}
\end{figure}

\subsection{Faster Learning Pace}
Controlling the effect of supervision to train neural networks not only improves the performance, but also provides the network with more solid signals which speeds up the learning process. 
Figure~\ref{fig:plot} illustrates the training/validation loss for both networks, compared to the loss of training the \tnet with weak supervision, along with their performance on test sets, with respect to different amounts of training data for the sentiment classification task\footnote{We have observed similar speed-up in the learning process of the ranking task, however we skip bringing its plots due to space limit since we have nested cross-validation for the ranking task and a set of plots for each fold.}.

As shown, in the training, the loss of the \tnet in our model, i.e. $\mathcal{L}_t$ is higher than the loss of the network which is trained only on weakly supervised data, i.e. $\mathcal{L}_\text{WSO}$. 
However, since these losses are calculated with respect to the weak labels (not true labels), having very low training loss can be an indication of overfitting to the imperfection in the weak labels. 
In other words, regardless of the general problem of lack of generalization due to overfitting, in the setup of learning from weak labels, predicting labels that are similar to train labels (very low training loss) is not necessarily a desirable incident. 

In the validation set, however, $\mathcal{L}_t$ decreases faster than $\mathcal{L}_\text{WSO}$, which supports the fact that $\mathcal{L}_\text{WSO}$ overfits to the imperfection of weak labels, while our setup helps the \tnet to escape from this imperfection and do a good job on the validation set.
In terms of the performance, compared to WSO, the performance of CWS on both test sets increases very quickly and CWS is able to pass the performance of the \wa by seeing much fewer instances annotated by the \wa.

\section{Related Work}
Learning from weak or noisy labels has been studied in the literature~\cite{Frenay:2014}. We briefly review research most relevant to our work.

\mypar{Semi-supervised learning.}
There are semi-supervised learning algorithms~\cite{Zhu:2005} developed to utilize weakly or even unlabeled data. Self-training~\cite{Rosenberg:2005} or pseudo-labeling~\cite{Lee:2013} tries to predict labels of unlabeled data. This unlabeled data is provided additionally. In particular for neural networks, methods use greedy layer-wise pre-training of weights using unlabeled data alone followed by supervised fine-tuning~\cite{Deriu:2017,Severyn:2015:SemEval,Severyn:2015:SIGIR,Go:2009}. Other methods learn unsupervised encodings at multiple levels of the architecture jointly with a supervised signal~\cite{Ororbia:2015,Weston:2012}.

\mypar{Meta-learning.} 
From the meta-learning perspective, our approach is similar to \citet{Andrychowicz:2016} where a separate recurrent neural network called \textit{optimizer} learns to predict an optimal update rule for updating parameters of the \tnet. The optimizer receives a gradient from the \tnet and outputs the adjusted gradient matrix. As the number of parameters in modern neural networks is typically on the order of millions the gradient matrix becomes too large to feed into the optimizer, so the approach of \citet{Andrychowicz:2016} is applied to very small models. In contrast, our approach leverages additional weakly labeled data where we use the \cnet to predict per-instance scores that calibrate gradient updates for the \tnet.

\mypar{Direct learning with weak/noisy labels.}
Many studies tried to address learning in the condition of imperfect labels. 
Some noise cleansing methods were proposed to remove or correct mislabeled instances~\cite{Brodley:1999}.
Other studies showed that weak or noisy labels can be leveraged by employing a particular architecture or defining a proper loss function to avoid overfitting the training data imperfection~\cite{Dehghani:2017:SIGIR, Patrini:2016, Beigman:2009, Zeng:2015, Bunescu:2007}.  

\mypar{Modeling imperfection.}
There is also research trying to model the pattern of the noise or weakness in the labels. 
Some methods leverage generative models to denoise weak supervision sources that a discriminative model can learn from~\cite{Ratner:2016,Rekatsinas:2017,Varma:2017}.
Other methods aim to capture the pattern of the noise by inserting an extra layer or a separated module~\cite{Sukhbaatar:2014,Veit:2017}, infer better labels from noisy labels and use them to supervise the training of the network.
This is inspired by the teacher-student paradigm~\cite{Hinton:2015,Romero:2014,Xiao:2015} in which the teacher generates a new label given the training instance with its corresponding weak or noisy label.
However, as we show in our experiments, this approach is not sufficient when the amount of supervised data is not enough to generate better labels. 

\section{Conclusion and Future Directions}
Training neural networks using large amounts of weakly annotated data is an attractive approach in scenarios where an adequate amount of data with true labels is not available.
In this paper, we propose a  multi-task neural network architecture that unifies learning to estimate the confidence score of weak annotations and training neural networks to learn a target task with controlled weak supervision, i.e. using weak labels to updating the parameters, but taking their estimated confidence scores into account. This helps to alleviate updates from instances with unreliable labels that may harm the performance.

We applied the model to two tasks, document ranking and sentiment classification, and empirically verified that the proposed model speeds up the training process and obtains more accurate results. 
As a promising future direction, we are going to understand to which extent using weak annotations has the potential of training high-quality models with neural networks and understand the exact conditions under which our proposed method works.

\bibliography{ref}
\bibliographystyle{emnlp_natbib}

\end{document}